\documentclass[letterpaper, 10 pt, journal, twoside]{IEEEtran}  %
\usepackage{graphicx}
\usepackage{times}
\usepackage{mathtools}
\usepackage{tabularx}
\usepackage{caption}
\usepackage{amsmath, amssymb}
\usepackage{url}
\usepackage{bm}
\usepackage[table, dvipsnames]{xcolor}
\definecolor{mySourceColor}{RGB}{216, 80, 63}
\definecolor{myTargetColor}{RGB}{91, 143, 229}
\definecolor{Vermilion}{HTML}{D55E00} 
\definecolor{MyMagenta}{RGB}{230, 0, 172}
\definecolor{linkpink}{HTML}{E6007E}

\usepackage{cite}
\usepackage{booktabs}

\usepackage{multirow}
\usepackage{enumitem}
\definecolor{ourscellbg}{RGB}{230,240,250}
\usepackage{hyperref}
\usepackage[normalem]{ulem}
\usepackage{makecell}

\usepackage[most]{tcolorbox}
\tcbuselibrary{theorems}

\newcommand{\meanstd}[2]{#1{\tiny\textcolor{gray}{$\pm$#2}}}
\newcommand{\bmeanstd}[2]{\textbf{#1}{\tiny\textcolor{gray}{$\pm$#2}}}
\newcommand{\umeanstd}[2]{\uline{#1}{\tiny\textcolor{gray}{$\pm$#2}}}

\title{
BooST: Bridging Semantics and Motions for Efficient Skill Transfer
}

\author{Jusuk Lee$^{1}$, Daesol Cho$^{2}$, Jonghun Shin$^{1}$, Seungyeon Yoo$^{1}$, Jonghae Park$^{1}$, Taekbeom Lee$^{1}$, and H. Jin Kim$^{1}$%
\thanks{Manuscript received: December 2, 2025; Revised July 12, 2026; Accepted August 8, 2026. This paper was recommended for publication by Editor Jens Kober upon evaluation of the Associate Editor and Reviewers' comments. This work was supported by Samsung Research Funding \& Incubation Center of Samsung Electronics under Project Number SRFC-IT2402-17}%
\thanks{$^{1}$ Jusuk Lee, Jonghun Shin, Seungyeon Yoo, Jonghae Park, Taekbeom Lee, H. Jin Kim are with Department of Aerospace Engineering, Automation and Systems Research Institute (ASRI), Seoul National University, Seoul, Korea {\tt\footnotesize \{jlee24, tlswhdgns0406, syeon.yoo, bdfire1234, ltb1128, hjinkim\}@snu.ac.kr}}%
\thanks{$^{2}$Daesol Cho is with Georgia Institute of Technology, Atlanta, GA, USA {\tt\footnotesize chodaesol@gmail.com}}
\thanks{Digital Object Identifier (DOI): see top of this page.}
\\[2mm]
\normalfont\normalsize\href{https://boost-robots.github.io/}{\textcolor{linkpink}{\texttt{\textbf{https://boost-robots.github.io}}}}
}

\IEEEaftertitletext{\vspace{-1\baselineskip}}
\begin{document}

\maketitle

\begin{abstract}
Skill abstraction---the process of learning reusable and temporally extended behaviors---has emerged as a key paradigm for improving sample efficiency and generalization in robot learning. For efficient skill transfer to real robots, learned skills must generalize across tasks and domains, remain robust to visual and dynamic perturbations, and be efficient enough for practical deployment. However, existing methods typically satisfy only a subset of these properties, as they capture either high-level semantic intent (\emph{what}) or low-level motion dynamics (\emph{how}). This incomplete skill transfer yields weak priors for policy learning, thereby demanding substantial in-domain data for downstream adaptation. To address these challenges, we introduce BooST, a two-stage framework that explicitly bridges semantics and motions to satisfy all three desiderata. BooST first leverages a cross-modal VQ-VAE to capture both semantic intent and motion dynamics, yielding a unified skill representation. It then distills this representation into a lightweight policy for efficient downstream adaptation to new tasks. Extensive experiments across simulation and real-robot settings demonstrate that BooST achieves superior few-shot adaptation, cross-domain skill transfer, and robustness to dynamic visual distractors, while maintaining a lightweight yet expressive design suitable for real-world deployment.
\end{abstract}


\section{Introduction}
\IEEEPARstart{A}{ central} objective in robot learning is to develop general-purpose agents capable of rapidly adapting to novel tasks in a zero-shot or few-shot manner. One promising direction toward this goal is large-scale pretraining of reusable robotic representations, inspired by the success of foundation models in computer vision and natural language processing~\cite{kirillov2023segment, oquab2024dinov2}. In this context, skill abstraction has emerged as an effective paradigm for learning such robotic representations. By offering compact and reusable representations of temporally extended behaviors that capture task-agnostic structures, skills serve as a versatile prior that enables efficient adaptation in the low-data regime of robot learning~\cite{mete2024quest, garg2022lisa}. Therefore, transferring skills learned from large-scale pretraining enables agents to efficiently adapt to novel tasks, even under limited data availability.

Achieving such efficient skill transfer to real robots requires three key properties: \emph{generalization}, \emph{robustness}, \emph{efficiency}. Skill representations should exhibit strong \emph{generalization}, enabling transfer across novel scenes, tasks, and even to different robotic embodiments. Realizing this level of generalization requires models to leverage extensive and heterogeneous datasets, which inevitably contain visual noise~\cite{ye2024latent, bu2025univla, kim2025uniskill}. Accordingly, skill learning should demonstrate \emph{robustness} to visual distractors and background variations, allowing effective pretraining without relying on filtered or curated data. Moreover, practical skill transfer requires \emph{efficiency}, entailing lightweight yet expressive policies for real-world deployment. While strong generalization typically demands large-scale models with high representational capacity~\cite{oquab2024dinov2, ye2024latent}, such models are computationally prohibitive for real-world robots due to latency and resource constraints.

\begin{figure}[!t]
  \centering
  \includegraphics[width=1.0\columnwidth]{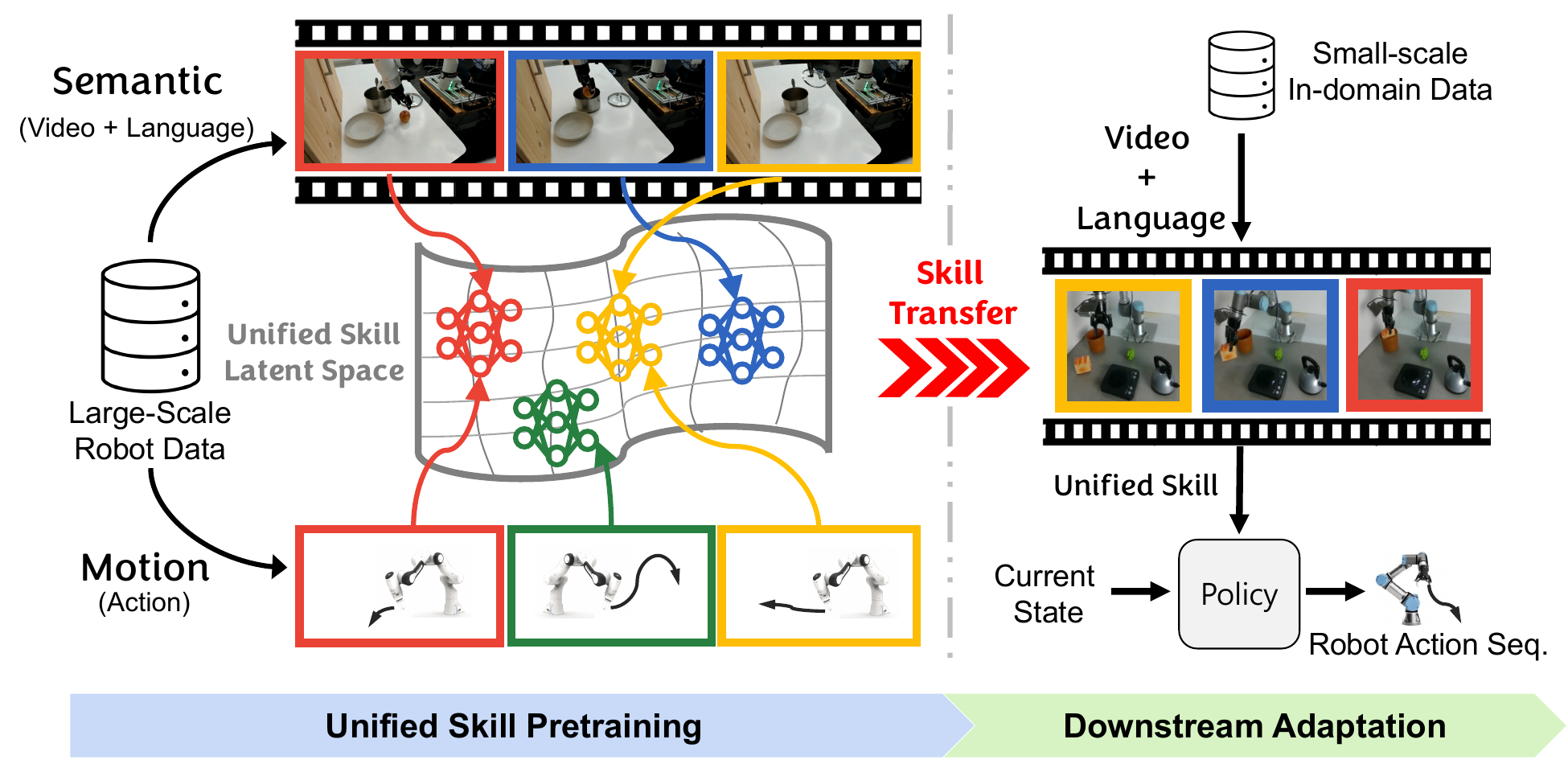}
  \caption{\textbf{Overview of BooST.} Our framework consists of two stages: (left) Unified Skill Pretraining, which learns a cross-modal skill representation from large datasets, and (right) Downstream Adaptation, which distills the learned skills into a lightweight policy and skill prior for efficient adaptation.}
  \label{fig:Thumbnail}
  \vspace{-5mm}
\end{figure}

To satisfy these properties, existing methods have learned skills by abstracting information at different levels: low-level methods abstract from action data, while high-level methods abstract from visuo-linguistic data. Low-level methods~\cite{lee2024behavior, mete2024quest, pertsch2021accelerating, zheng2024prise} focus on the \emph{how}, modeling the fine-grained motion dynamics across diverse trajectories. However, their reliance on embodiment-specific behaviors and the lack of semantic grounding jointly hinder \emph{generalization} across robotic embodiments and tasks.
High-level methods~\cite{garg2022lisa, liang2024skilldiffuser, jiangdiscrete, zhang2024extract, wan2024lotus, ye2024latent, bu2025univla} focus on the \emph{what}, capturing the semantic intent inferred from an image or language instruction (e.g., pick up things), enabling cross-domain transfer.
Yet their dependence only on visuo-linguistic features makes them sensitive to visual distractors, undermining \emph{robustness}. Moreover, several high-level approaches~\cite{ye2024latent, bu2025univla} rely on large-scale models to compensate for insufficient guidance from learned skills to the policy, which in turn reduces \emph{efficiency}.
Ultimately, both directions fall short because they capture only one aspect of a skill—either the motion dynamics or the semantic intent. As a result, their learned representations fail to transfer in a way that achieves generalization, robustness, and efficiency simultaneously, requiring substantial in-domain data for adaptation to new scenes and tasks.

In this paper, we introduce BooST (\textbf{B}ridging semantics and m\textbf{o}ti\textbf{o}ns for Efficient \textbf{S}kill \textbf{T}ransfer), a framework for learning a unified skill representation that captures both semantic intent (\emph{what}) and motion dynamics (\emph{how}). 
This representation enables cross-domain transfer with practical deployability and robustness to visual distractors, thereby facilitating few-shot adaptation.
BooST adopts a decoupled two-stage training paradigm. In the \textbf{first stage}, it pretrains a unified skill representation on large-scale and diverse datasets to jointly encode high-level semantics and low-level motions. At its core, BooST employs a cross-modal VQ-VAE~\cite{yoo2024mono, spurr2018cross} that learns a shared codebook via two pathways. The visuo-linguistic pathway extracts semantic intent from visual observations and language instructions using a pretrained CLIP model~\cite{radford2021learning} and a cross-attention mechanism. In parallel, the action pathway encodes motion dynamics from action trajectories, grounding the learned semantics in executable behaviors. The proposed design also underpins BooST's robustness to dynamic visual distractors (e.g., irrelevant moving objects) and background variations (e.g., changes in surrounding layouts and objects). The resulting skill representation generalizes effectively across diverse scenes, tasks, and robotic embodiments, yielding highly transferable skills. In the \textbf{second stage}, this rich representation is distilled into a lightweight policy and skill prior, enabling efficient downstream adaptation.

We validate BooST through both simulation and real-world robot experiments, focusing on few-shot adaptation. Across various data regimes, BooST consistently outperforms existing skill-based methods, particularly in the low-data regime. Notably, in real-world settings, BooST achieves multi-task adaptation with only five demonstrations per task, and further generalizes across new scenes, tasks, and even different robotic embodiments.
In addition, BooST demonstrates robust skill learning when pretrained in environments containing dynamic visual distractors, highlighting its resilience to irrelevant visual noise. 

To summarize, our main contributions are:
\begin{itemize}
  \item We present BooST, a framework that learns a unified skill representation by integrating semantic intent and motion dynamics into a shared skill codebook.
  \item We introduce a decoupled two-stage training pipeline that reconciles large-scale pretraining with deployment efficiency, enabling efficient adaptation of a lightweight policy to new tasks in a few-shot manner.
  \item We demonstrate that BooST achieves strong few-shot adaptation, cross-embodiment transfer, and robustness to dynamic visual distractors and background variations across simulation and real-world experiments.
\end{itemize}

\section{Related Works}
\subsection{Skill Abstraction from Offline Data}
\paragraph{Low-level skill abstractions}
Low-level skill methods~\cite{lee2024behavior, mete2024quest, pertsch2021accelerating, zheng2024prise} quantize action trajectories into discrete latent skill spaces using residual~\cite{zeghidour2021soundstream} or finite scalar vector quantization~\cite{mentzer2023finite}. Relying solely on action sequences, these methods cannot infer which action to perform from visual or language inputs. This lack of semantic grounding degrades downstream performance~\cite{nair2022r3m, ma2023liv} and increases the data required for adaptation. Moreover, low-level skills are tightly coupled to the pretraining action space, so skill transfer succeeds only when the downstream environment shares the same action space. For example, skills learned from joint velocity action space do not transfer to downstream tasks that use Cartesian end-effector position control. In contrast, BooST integrates semantic understanding with motion dynamics, yielding semantically grounded skills. Furthermore, BooST leverages the visuo-linguistic pathway alone during skill transfer, disentangling the learned representation from the pretraining action space. This design enables skill transfer across heterogeneous action spaces, supporting cross-embodiment skill transfer.
\paragraph{High-level skill abstractions}
High-level skill methods map visual and linguistic inputs into discrete latent skill spaces to encode semantic intent. One line of work~\cite{garg2022lisa, jiangdiscrete, liang2024skilldiffuser} quantizes past observations and language instructions into discrete skills through vector quantization~\cite{gray1984vector}, but trains visual encoder from scratch without pretrained foundation models. As a result, the learned skills overfit visual patterns in the training data, become sensitive to minor appearance changes, and transfer poorly to novel scenes and tasks. More recent methods~\cite{zhang2024extract, wan2024lotus} leverage vision foundation models~\cite{oquab2024dinov2, nair2022r3m} to improve generalization, but define skills primarily in terms of visual changes without grounding them in actions or language. Their representations thus remain vulnerable to visual distractors and fail to capture meaningful motion dynamics. BooST instead fuses image and language features from a pretrained CLIP model via cross-attention to obtain task-aware features. Together with the action reconstruction objective and the explicit action pathway, this design yields motion-grounded skills that are robust to dynamic visual distractors and background variations.

\subsection{Latent Action Pretraining}
\begin{figure*}[t!]
\centering
\includegraphics[width=1.0\textwidth]{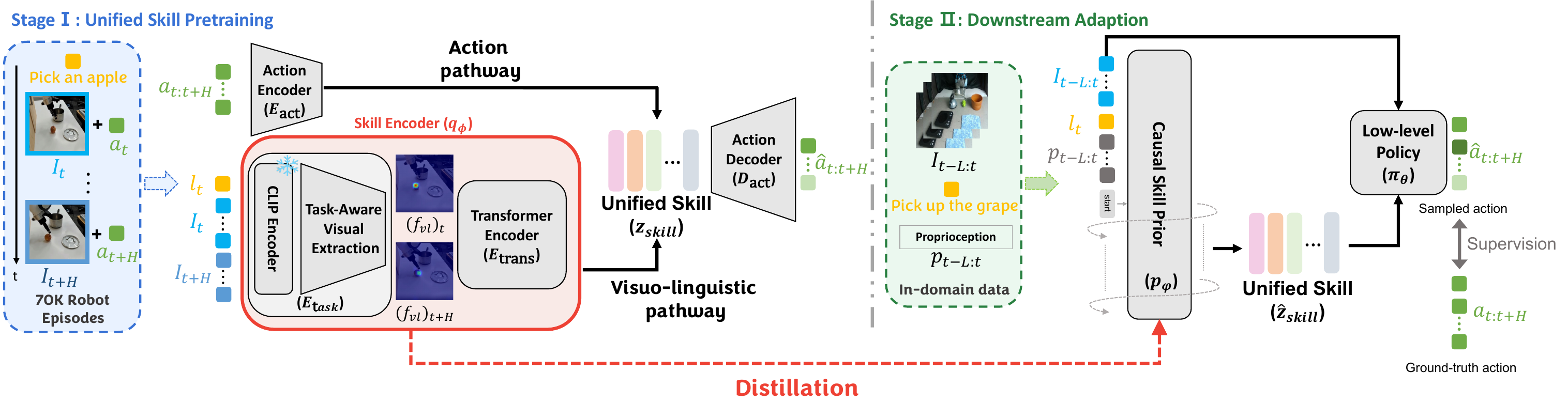}
\caption{The framework consists of two stages: (Left) \textbf{Stage I: Unified Skill Pretraining}, which learns a cross-modal unified skill representation by jointly encoding semantic intent (visuo-linguistic pathway) and motion dynamics (action pathway); and (Right) \textbf{Stage II: Downstream Adaptation}, which distills the learned skill representation into a lightweight causal skill prior and low-level policy for efficient adaptation to new tasks. Attention heatmaps within the skill encoder indicate that BooST focuses on task-relevant regions during pretraining. The attention heatmaps are best viewed in the digital version.}
\label{fig:main_figure}
\vspace{-4mm}
\end{figure*}

Recent approaches learn latent actions that describe what motion follows from visual and linguistic inputs~\cite{bruce2024genie, ye2024latent, kim2025uniskill, bu2025univla}. These methods use large-scale human or robot video datasets and apply vector quantization~\cite{gray1984vector} to encode visual changes between frames into discrete latent spaces. The resulting representations have been primarily used to pretrain vision-language-action (VLA) models for downstream policy learning. However, because these approaches define latent actions purely from visual differences, the learned spaces tend to capture task-irrelevant motion when dynamic visual distractors are present, degrading policy performance. In contrast, BooST learns discrete latent skills that remain robust under dynamic visual distractors by jointly leveraging an action pathway, explicit action supervision, and task-aware visual features. Furthermore, our method requires no manual curation of robot data, enabling the use of easy-to-access and continually growing in-the-wild robot datasets~\cite{khazatsky2024droid}.

\section{Preliminaries}
\subsection{Background}
We briefly review the three components of BooST. VQ-VAE and its variants~\cite{vali2022nsvq, zeghidour2021soundstream} are an autoencoder that learns a discrete latent representation. An encoder maps an input to a continuous vector, a learnable codebook quantizes it to the nearest code, and a decoder reconstructs the input, with all three trained jointly by a reconstruction objective. The CLIP image encoder~\cite{radford2021learning} is a Vision Transformer trained with contrastive image--language pretraining, producing semantic image features aligned with language. BAKU~\cite{haldar2024baku} is a transformer-based policy that fuses multimodal observations and predicts action trajectories.

\subsection{Problem Setup}
Our goal is to learn a policy that enables a robot to perform a wide range of tasks specified by natural language instructions. We assume access to a dataset of \(N\) language‐conditioned trajectories \(\mathcal{D} = \bigl\{\bigl(l^{(i)},\,O^{(i)}_1,a^{(i)}_1,\dots,O^{(i)}_{T_i},a^{(i)}_{T_i}\bigr)\bigr\}_{i=1}^N\), where each trajectory is paired with a natural language instruction \(l\), and each observation \(O_t = (I_t, I_t^{\text{gripper}}, p_t)\) consists of a front camera image, a gripper camera image (if available), and the robot’s proprioceptive state, while the action \(a_t\) is continuous-valued. Our objective is to learn a language-conditioned policy \(\Pi(a_{t:t+H} \mid O_{t-L:t}, l)\). Here, \(L\) denotes the observation history length and \(H\) is the action prediction horizon.

\section{Methods}
Our method decouples large-scale skill pretraining from small-scale downstream adaptation, as illustrated in Fig.~\ref{fig:main_figure}.
Section~\ref{sec:method_decoupling} introduces the overall learning formulation and explains how the training objective derived from the Evidence Lower Bound (ELBO) naturally motivates this two-stage design. Section~\ref{sec:method_pretraining} describes the large-scale skill pretraining phase, where a unified skill representation is learned from diverse offline datasets. Section~\ref{sec:method_downstream} presents the downstream adaptation phase, in which a lightweight skill prior and policy are efficiently trained for target tasks. 

\subsection{Decoupling Skill Learning and Downstream Adaptation}
\label{sec:method_decoupling}
Training a monolithic, large-scale policy \(\Pi\) yields powerful but computationally impractical models for real-world deployment. To enable lightweight yet expressive policy learning, we reformulate the objective of \(\Pi\) within a variational framework by introducing a latent skill variable \(z\). 
Marginalizing over \(z\) decomposes \(\Pi\) into a low-level policy \(\pi_\theta\) and a skill prior \(p\):
\begin{equation}
\begin{aligned}
&\log \Pi(a_{t:t+H} \mid O_{t-L:t}, l) \\
&= \log \sum_{z_t} \pi_{\theta}(a_{t:t+H} \mid z_t, O_{t-L:t}, l)\, p(z_t \mid O_{t-L:t}, l).
\end{aligned}
\end{equation}
Rewriting the marginal as an expectation under a skill encoder \(q_\phi\) and applying Jensen's inequality yields the ELBO:
\begin{multline}
\mathcal{L}(\theta, \psi, \phi)
= \mathbb{E}_{z_t \sim q_{\phi}}[\log \pi_{\theta}(a_{t:t+H} \mid z_t, O_{t-L:t})] \\
- D_{\mathrm{KL}}(q_{\phi}(z_t\mid I_t, I_{t+H}, l)
\parallel p_{\psi}(z_t \mid O_{t-L:t}, l)).
\end{multline}

The resulting ELBO decomposes the learning problem into a skill encoder \(q_{\phi}\), a skill prior \(p_{\psi}\), and a low-level policy \(\pi_{\theta}\), and maps directly to our training objectives: the KL term becomes the prior distillation loss for \(p_{\psi}\), and the action-likelihood term becomes the behavior-cloning loss for \(\pi_{\theta}\) (Eq.~\ref{eq:downstream_loss_sum}).
To distill the expressiveness of large-scale models into compact policies, we adopt a two-stage training framework: \textbf{Stage I} pretrains the skill encoder \(q_{\phi}\) on large-scale datasets to learn a unified skill representation, and \textbf{Stage II} distills this representation into the lightweight prior \(p_{\psi}\) and policy \(\pi_{\theta}\) through downstream adaptation. This decoupled, distillation-based approach preserves the representational richness of large models while maintaining computational efficiency for real-world robotic deployment.

\subsection{Stage I: Unified Skill Pretraining}
\label{sec:method_pretraining}
The goal of Stage I is to pretrain a unified skill representation from large-scale, diverse offline data that jointly captures high-level semantic intent (\emph{what}) and low-level motion dynamics (\emph{how}). This stage, illustrated on the left side of Fig.~\ref{fig:main_figure}, corresponds to the unified skill pretraining phase of BooST. To this end, we adopt a cross-modal VQ-VAE framework that learns a discrete latent space of reusable skills. Two complementary pathways contribute to the skill learning process: a visuo-linguistic pathway for semantic grounding and an action pathway for motion dynamics encoding. We alternately optimize these two pathways to ensure a balanced and robust skill representation.

\begin{figure}[!t]
  \centering 
  \includegraphics[width=1.0\columnwidth]{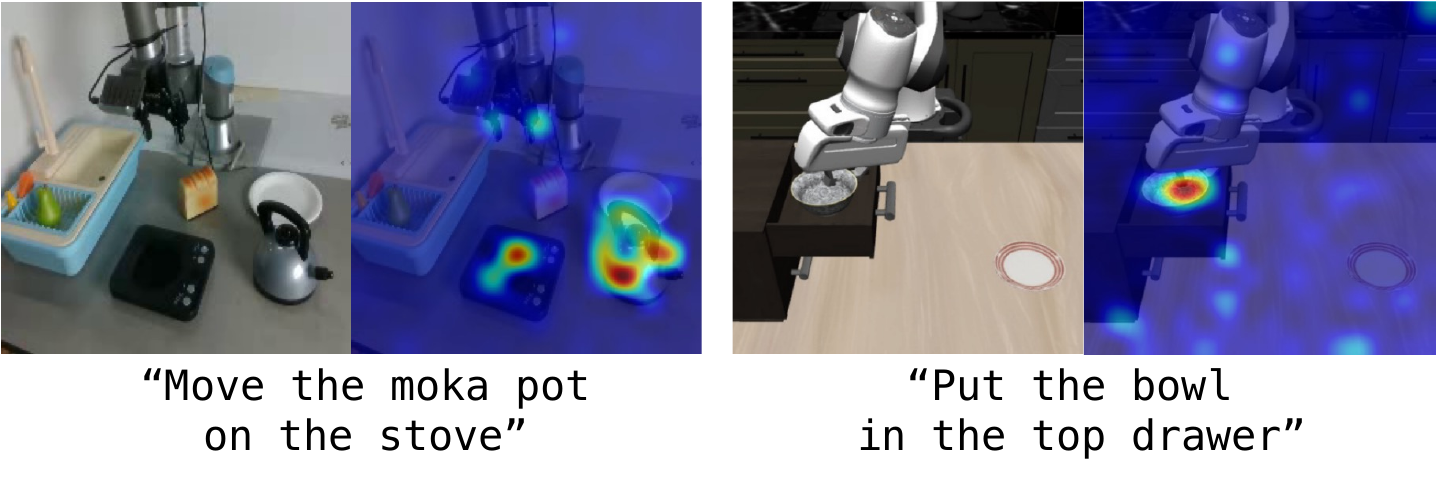}
  \caption{Attention maps from the visuo-linguistic pathway showing that BooST attends to instruction-relevant regions conditioned on the task instruction.}
  \label{fig:attention_map_example}
  \vspace{-6mm}
\end{figure}

The visuo-linguistic pathway provides semantic context through our skill encoder \(q_{\phi}\), defined as \(q_{\phi}(\cdot) = E_{\text{trans}}(E_{\text{task}}(\cdot))\). Inspired by recent works~\cite{lan2024clearclip, huang2025otter}, the task-aware encoder \(E_{\text{task}}\) takes the current and future images \((I_t, I_{t+H})\) along with the language instruction \(l\), and fuse patch-level visual tokens from a pretrained CLIP ViT with the instruction embedding through temperature-scaled cross-attention:
\begin{equation}
  \centering
  \label{eq:text_aware_visual_feature}
  f_{\mathrm{vl}} = \mathrm{softmax}\!\Bigl(\tfrac{\hat f_l\,\hat f_v^\top}{\tau}\Bigr)\,(\hat f_v + \mathrm{PE}),
\end{equation}
where \(\hat f_v\) and \(\hat f_l\) denote normalized visual and language features, respectively, \(\tau\) is a learnable temperature, and PE denotes positional embeddings. This fusion allows the model to selectively attend to instruction-relevant visual regions (see Fig.~\ref{fig:attention_map_example} for qualitative attention visualization). The fused features are subsequently processed by a transformer encoder \(E_{\text{trans}}\), which models temporal relationships between the current and future frames, producing the final visuo-linguistic embedding \(f_{\text{enc},\mathrm{vl}}= E_{\text{trans}}((f_{\mathrm{vl}})_t, (f_{\mathrm{vl}})_{t+H})\). In parallel, the action pathway encoder \(E_{\text{act}}\) processes action trajectories to capture fine-grained motion dynamics, producing dynamic features \(f_{\text{enc, act}}=E_{\text{act}}(a_{t:t+H})\). Both pathways focus on extracting task-relevant information, which is subsequently quantized into a shared discrete skill codebook. The continuous feature vector \(f_{\text{enc}}\) from either pathway (i.e., \(f_{\text{enc},\mathrm{vl}}\) or \(f_{\text{enc, act}}\)) is then quantized to the nearest vector \(c_k\) in the codebook, \(\mathcal{C} = \{c_k \}_{k=1}^K\), to obtain the discrete skill \(z_t\):
\begin{equation}
\label{eq:quantization}
z_t
= \operatorname*{arg\,min}_{k\in\{1,\dots,K\}}
  \bigl\lVert f_{\mathrm{enc}} - c_k\bigr\rVert_2^2.
\end{equation}
The framework is optimized via a single supervisory signal, which is action reconstruction. The action decoder \(D_{\text{act}}\) reconstructs action sequences from the discrete skill \(z_t\) and current task-aware visual feature \((f_{\mathrm{vl}})_t\), i.e., \(\hat{a}_{t:t+H} = D_{\text{act}}(z_t, (f_{\mathrm{vl}})_t)\). Unlike prior approaches that reconstructs pixel-level images~\cite{ye2024latent, kim2025uniskill, bu2025univla}, BooST employs low-dimensional action sequences as the reconstruction target. This design encourages the model to ignore visual details irrelevant to the task, and the action pathway further reinforces robustness against dynamic visual distractors. We employ residual vector quantization~\cite{zeghidour2021soundstream} and apply a rotation trick~\cite{fifty2024restructuring} to prevent codebook collapse. The overall pretraining objective is defined as a weighted sum of reconstruction losses from both pathways:
\begin{equation}
\begin{split}
    \mathcal{L}_{\text{pretrain}}(\phi) ={} 
    & \lambda_1 \underbrace{\lVert a_{t:t+H} - \hat{a}_{\mathrm{vl}} \rVert_2^2}_{\substack{\text{visuo-linguistic} \\ \text{pathway}}}
    + \lambda_2 \underbrace{\lVert a_{t:t+H} - \hat{a}_{\text{act}} \rVert_2^2}_{\substack{\text{action} \\ \text{pathway}}} \\
    & + \lambda_3 \underbrace{\lVert f_{\mathrm{enc, vl}} - f_{\mathrm{enc, act}} \rVert_2^2}_{\text{regularization}}
\end{split}
\end{equation}
where \(a_{t:t+H}\) is the ground-truth action sequence, and \(\hat{a}_{\mathrm{vl}}\) and \(\hat{a}_{\text{act}}\) are the reconstructions from the two pathways. 
In our experiments, we set \(\lambda_1 = 3\), \(\lambda_2 = 1\), and \(\lambda_3 = 0.5\), and use a residual vector quantization codebook of size \(K = 16\) with \(2\) quantization levels.

\subsection{Stage II: Downstream Adaptation}
\label{sec:method_downstream}
In Stage II, we distill the rich knowledge encoded in the pretrained skill encoder \(q_{\phi}\) into two lightweight components: a skill prior \(p_{\psi}\) and a low-level policy \(\pi_{\theta}\), both designed for efficient deployment. This stage, illustrated on the right side of Fig.~\ref{fig:main_figure}, corresponds to the downstream adaptation phase of BooST. Unlike Stage I, which leverages future image to learn predictive representations, Stage II uses only past observations, as future inputs are unavailable during execution. Both \(p_{\psi}\) and \(\pi_{\theta}\) are trained on a smaller, in-domain dataset and implemented as small Transformer models that process multimodal input stream consisting of camera features and proprioceptive states. The policy head follows BAKU~\cite{haldar2024baku} and models the action distribution as an isotropic Gaussian.

\begin{multline}
\label{eq:downstream_loss_sum}
\mathcal{L}_{\text{downstream}}(\theta,\psi;\phi)
= 
\underbrace{\mathbb{E}_{z_t^q\sim q_{\phi}}
\bigl[-\log p_{\psi}(z_t^q \mid O_{t-L:t},l)\bigr]}_{\text{Prior Distillation Loss}} \\
+\, \alpha
\underbrace{\mathbb{E}_{z_t\sim p_\psi}
\Bigl[-\log \pi_{\theta}(a_{t:t+H}\mid \operatorname{sg}(z_t),O_{t-L:t})\Bigr]}_{\text{Policy Behavior Cloning Loss}}.
\end{multline}

The first term trains the skill prior \(p_{\psi}\) to approximate the distribution of latent skills predicted by the frozen skill encoder \(q_{\phi}\), where \(z_t^q \sim q_{\phi}(I_t, I_{t+H}, l)\) serves as a pseudo-label. The second term optimizes the policy \(\pi_{\theta}\) to imitate expert actions conditioned on sampled skills. For stable training, the stop-gradient operator \(\operatorname{sg}\) prevents gradient flow from the policy into the prior, ensuring that the prior is supervised solely by the fixed encoder outputs.

\section{Experiments}
\begin{table*}[t]
  \centering
  \caption{\textbf{Few-shot adaptation performance on LIBERO benchmarks.}
  We report the mean success rate with 95\% confidence intervals over five random seeds, each evaluated across 50 rollouts per task.
  The downstream skill prior and policy are trained with varying numbers of demonstrations (50, 20, and 10).
  \textbf{Bold} and \underline{underlined} numbers indicate the best and the second-best mean in each column, respectively.
  Relative improvements (\textcolor{MyMagenta}{\scriptsize\texttt{+X\%}}) are computed over the second-best method. This relative improvement generally increases as the number of demonstrations decreases (from 50 down to 10).}
  \label{tab:simulation_main}
  \scriptsize
  \renewcommand{\arraystretch}{1.2}
  \setlength{\tabcolsep}{3pt}
  \begin{tabularx}{\linewidth}{%
      l 
      *{3}{>{\centering\arraybackslash}X}
      *{3}{>{\centering\arraybackslash}X}
      *{3}{>{\centering\arraybackslash}X}
      *{3}{>{\centering\arraybackslash}X}
    }
    \toprule
    Method
      & \multicolumn{3}{c}{\shortstack{LIBERO-90\\(\# demos)}}
      & \multicolumn{3}{c}{\shortstack{LIBERO-Goal\\(\# demos)}}
      & \multicolumn{3}{c}{\shortstack{LIBERO-Object\\(\# demos)}}
      & \multicolumn{3}{c}{\shortstack{LIBERO-Spatial\\(\# demos)}} \\
    \cmidrule(lr){2-4}\cmidrule(lr){5-7}\cmidrule(lr){8-10}\cmidrule(lr){11-13}
      & 50 & 20 & 10 & 50 & 20 & 10 & 50 & 20 & 10 & 50 & 20 & 10 \\
    \midrule
    Diffusion Policy~\cite{chi2023diffusion}
      & \meanstd{0.60}{0.07} & \meanstd{0.33}{0.11} & \meanstd{0.24}{0.10}
      & \meanstd{0.56}{0.15} & \umeanstd{0.47}{0.05} & \umeanstd{0.41}{0.04}
      & \meanstd{0.36}{0.28} & \meanstd{0.25}{0.21} & \meanstd{0.20}{0.09}
      & \meanstd{0.74}{0.03} & \meanstd{0.50}{0.08} & \umeanstd{0.42}{0.06} \\
    VQ-BeT~\cite{lee2024behavior}
      & \umeanstd{0.64}{0.12} & \umeanstd{0.51}{0.05} & \umeanstd{0.29}{0.07}
      & \umeanstd{0.73}{0.09} & \meanstd{0.45}{0.08} & \meanstd{0.32}{0.09}
      & \meanstd{0.36}{0.17} & \meanstd{0.32}{0.18} & \meanstd{0.16}{0.20}
      & \umeanstd{0.80}{0.07} & \umeanstd{0.63}{0.06} & \meanstd{0.37}{0.09} \\
    QueST~\cite{mete2024quest}
      & \meanstd{0.51}{0.02} & \meanstd{0.37}{0.02} & \meanstd{0.28}{0.02}
      & \meanstd{0.30}{0.06} & \meanstd{0.25}{0.06} & \meanstd{0.22}{0.05}
      & \meanstd{0.11}{0.01} & \meanstd{0.06}{0.05} & \meanstd{0.08}{0.07}
      & \meanstd{0.28}{0.03} & \meanstd{0.21}{0.02} & \meanstd{0.10}{0.04} \\
    LISA~\cite{garg2022lisa}
      & \meanstd{0.00}{0.00} & \meanstd{0.00}{0.00} & \meanstd{0.00}{0.00}
      & \meanstd{0.00}{0.00} & \meanstd{0.00}{0.00} & \meanstd{0.00}{0.00}
      & \meanstd{0.00}{0.00} & \meanstd{0.00}{0.00} & \meanstd{0.00}{0.00}
      & \meanstd{0.00}{0.00} & \meanstd{0.00}{0.00} & \meanstd{0.00}{0.00} \\
    EXTRACT~\cite{zhang2024extract}
      & \meanstd{0.22}{0.02} & \meanstd{0.20}{0.02} & \meanstd{0.14}{0.03}
      & \meanstd{0.09}{0.03} & \meanstd{0.06}{0.03} & \meanstd{0.04}{0.02}
      & \umeanstd{0.88}{0.04} & \umeanstd{0.69}{0.09} & \umeanstd{0.51}{0.12}
      & \meanstd{0.75}{0.04} & \meanstd{0.56}{0.12} & \meanstd{0.29}{0.09} \\
    \rowcolor{ourscellbg}
    \textbf{BooST (Ours)}
      & \bmeanstd{0.91}{0.01} \textcolor{MyMagenta}{\scriptsize(+41\%)} & \bmeanstd{0.82}{0.03} \textcolor{MyMagenta}{\scriptsize(+59\%)} & \bmeanstd{0.70}{0.02} \textcolor{MyMagenta}{\scriptsize(+140\%)}
      & \bmeanstd{0.92}{0.02} \textcolor{MyMagenta}{\scriptsize(+25\%)} & \bmeanstd{0.81}{0.02} \textcolor{MyMagenta}{\scriptsize(+74\%)} & \bmeanstd{0.68}{0.04} \textcolor{MyMagenta}{\scriptsize(+65\%)}
      & \bmeanstd{0.95}{0.03} \textcolor{MyMagenta}{\scriptsize(+8\%)} & \bmeanstd{0.85}{0.17} \textcolor{MyMagenta}{\scriptsize(+24\%)} & \bmeanstd{0.80}{0.09} \textcolor{MyMagenta}{\scriptsize(+57\%)}
      & \bmeanstd{0.91}{0.05} \textcolor{MyMagenta}{\scriptsize(+13\%)} & \bmeanstd{0.80}{0.05} \textcolor{MyMagenta}{\scriptsize(+27\%)} & \bmeanstd{0.60}{0.07} \textcolor{MyMagenta}{\scriptsize(+43\%)} \\
    \bottomrule
  \end{tabularx}
  \vspace{-4mm}
\end{table*}

We evaluate BooST in both simulation (LIBERO~\cite{liu2023libero}) and real-world settings (UR3 robot with a Robotiq 2F-85 gripper). For both settings, we pretrain BooST's unified skill representation on the large-scale DROID dataset~\cite{khazatsky2024droid}, which contains 76k teleoperated trajectories collected with a Franka Emika Panda arm equipped with a Robotiq 2F-85 gripper. 
The action space of the DROID dataset is defined in joint velocity space, whereas all downstream environments (LIBERO and UR3) operate in Cartesian end-effector space. We also train and evaluate the robustness of BooST under dynamic visual distractors and perform ablation studies to investigate key design choices.
We aim to address the following key questions in our experiments:
\begin{enumerate}[label=\textbf{(Q\arabic*)}]
    \item \label{rq:what-how} Does bridging semantic intent (\emph{what}) and motion dynamics (\emph{how}) lead to sample-efficient downstream adaptation compared to prior skill-based approaches? 
    \item \label{rq:embodiment} Can BooST enable skill transfer beyond novel scenes and tasks to different robotic embodiments?
    \item \label{rq:robustness} Can BooST robustly learn skills in environments with dynamic visual distractors? 
\end{enumerate}

\subsection{\textbf{(Q1)}: Downstream Adaptation in Simulation}
\label{sec:experiment_simulation}

\noindent\textbf{Experiment setup.} We evaluate how the unified skill representation affects few-shot adaptation performance on the LIBERO benchmark~\cite{liu2023libero}, which includes LIBERO-90, LIBERO-Goal, LIBERO-Object, and LIBERO-Spatial. We train downstream policies with 50, 20, or 10 demonstrations per task to assess sample efficiency under varying data regimes. Notably, none of the LIBERO samples appear in the pretraining dataset, necessitating generalization over novel scenes and tasks in skill transfer.

\noindent\textbf{Baselines.} We compare BooST with five representative methods covering both low- and high-level skill abstractions:
\begin{itemize}
    \item \textbf{Diffusion Policy}~\cite{chi2023diffusion} directly maps observations and language instructions to action sequences without skill abstraction.
    \item \textbf{VQ-BeT}~\cite{lee2024behavior} and \textbf{QueST}~\cite{mete2024quest} represent low-level skill approaches that quantize action trajectories into discrete motion primitives but lack semantic grounding.
    \item \textbf{LISA}~\cite{garg2022lisa} and \textbf{EXTRACT}~\cite{zhang2024extract} exemplify high-level skill approaches that focus on semantic intent. LISA encodes past observations and language instructions into discrete skills, jointly optimizing skill selection and policy execution. EXTRACT defines skills as visual feature transitions using pretrained vision foundation models. For fair comparison, we adapt EXTRACT to the imitation learning setting with the same skill prior, low-level policy, and behavior cloning objective as BooST.
\end{itemize}

\noindent\textbf{Results.} {Table~\ref{tab:simulation_main} reports the quantitative results across four LIBERO benchmarks under varying numbers of demonstrations. BooST consistently outperforms all baselines in every setting. Notably, the relative improvement over the second-best method becomes more pronounced as the number of demonstrations decreases, demonstrating BooST's strong few-shot adaptation capability. For example, on LIBERO-90 (the most diverse benchmark), BooST achieves relative gains of \(+140\%\), \(+59\%\), and \(+41\%\) with 10, 20, and 50 demonstrations, respectively. We attribute this improvement to the unified skill representation, which jointly captures semantic intent and motion dynamics, enabling effective transfer to novel scenes and tasks. As a qualitative validation, Fig.~\ref{fig:experiment_main_figure} shows that BooST preserves consistent semantic intent and motion dynamics across environments, successfully transferring skills from the DROID source domain to the LIBERO targets.

In contrast, both low-level and high-level baselines underperform BooST, because each captures only one aspect---motion dynamics or semantic intent---leaving their skill representations incomplete. These limitations are exemplified by high-level methods such as EXTRACT. It achieves competitive results on simple benchmarks (LIBERO-Object, Spatial) but underperforms on LIBERO-90 and Goal, where diverse motion dynamics are required. This is because defining skills purely through visual feature variations neglects motion fidelity. Notably, LISA completely fails in downstream adaptation experiments. Its joint optimization of skill and policy causes severe training instability on large and diverse datasets, often resulting in codebook collapse. This provides empirical evidence for our design choice to decouple skill learning from downstream adaptation.

\subsection{\textbf{(Q2)}: Real-World Cross-Embodiment Transfer}
\label{sec:experiment_real}
\begin{figure*}[t]
  \centering 
  \includegraphics[width=1.0\textwidth]{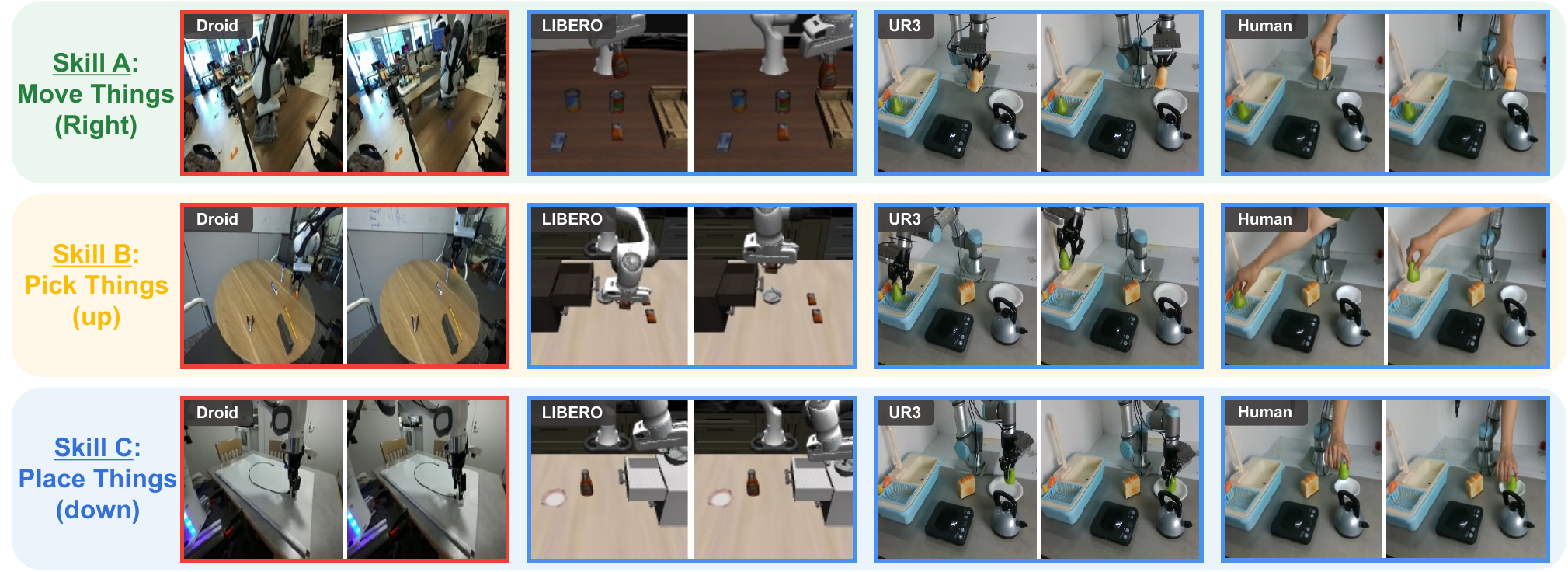}
  \caption{\textbf{Qualitative skill transfer across domains and embodiments.} Each row corresponds to a learned skill from the source domain (\textcolor{mySourceColor}{red}, Droid) and its execution in the target domains (\textcolor{myTargetColor}{blue}, LIBERO, UR3, and human hand). Each skill exhibits similar semantic intent and motion dynamics across distinct visual scenes and embodiments.}
  \label{fig:experiment_main_figure}
  \vspace{-5mm}
\end{figure*}

\noindent\textbf{Experiment setup.} We evaluate cross-embodiment skill transfer on a UR3 robot. The pretraining data was collected using a Franka Emika Panda arm, whereas the downstream experiments employ a UR3 robot with a different embodiment and action space. The setup includes two RGB cameras: a front-mounted and a wrist-mounted Intel RealSense D435i. We design four representative manipulation tasks and adapt policy using \emph{only five demonstrations}. The baseline methods are identical to those used in the simulation experiments. 

\noindent\textbf{Results.} The results in Fig.~\ref{fig:real_quantitative_results} clearly demonstrate that BooST enables sample-efficient skill transfer across heterogeneous robotic embodiments. Despite being pretrained on the Franka Emika Panda, BooST successfully transfers skills to the UR3 robot and achieves the highest success rates across all four real-world manipulation tasks. Remarkably, this performance is obtained with only five demonstrations per task, highlighting BooST's strong sample efficiency in real-world adaptation. As illustrated in Fig.~\ref{fig:experiment_main_figure}, the learned skills also preserve consistent semantic intent and motion dynamics across embodiments. In contrast, low-level methods such as VQ-BeT and QueST fail to transfer across embodiments because their skill representations are tied to the pretrained action space---e.g., skills learned from joint-velocity trajectories cannot generalize to robots controlled in Cartesian end-effector space.

\subsection{\textbf{(Q3)}: Robustness to Dynamic Visual Distractors}
\label{sec:experiment_robust}

\begin{figure*}[t]
  \centering 
  \includegraphics[width=0.9\textwidth]{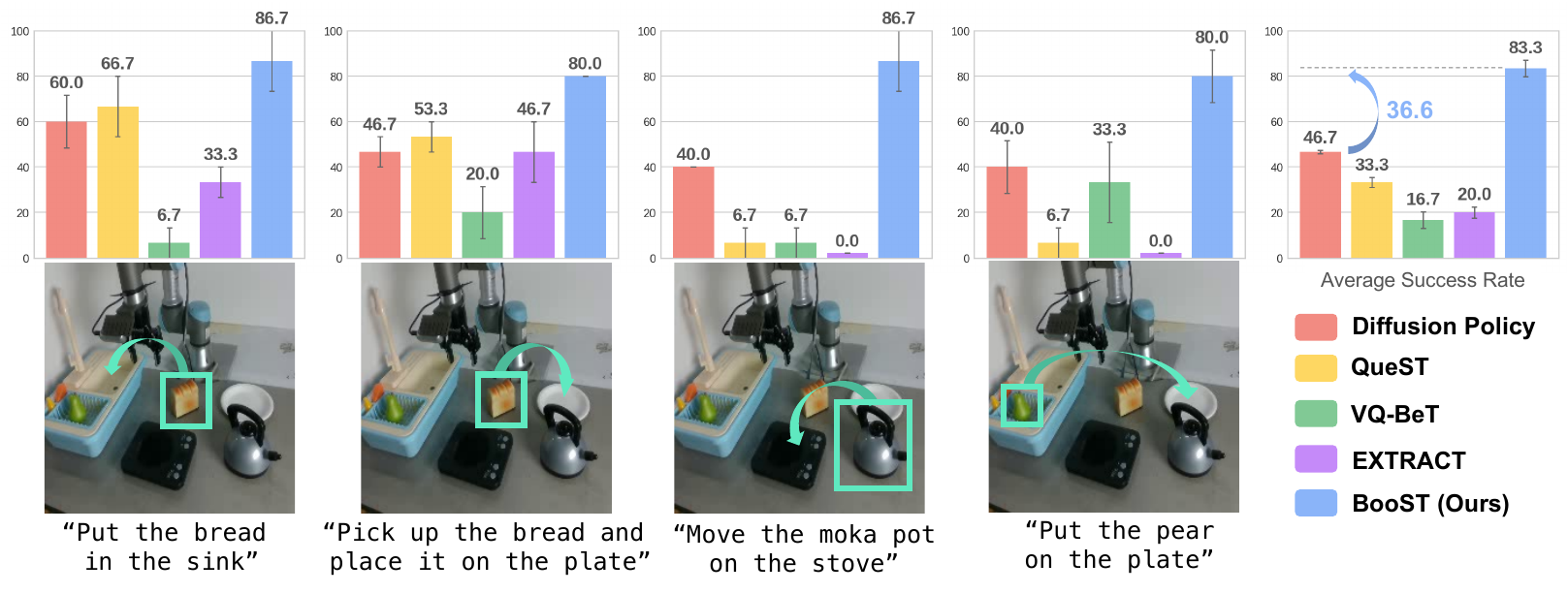}
  \caption{\textbf{Quantitative results on the UR3 robot.} Bar charts report average success rates over three random seeds (five evaluation trials per seed), with each method trained on only five demonstrations per task.}
  \label{fig:real_quantitative_results}
  \vspace{-5mm}
\end{figure*}

\noindent\textbf{Experiment setup.} We further evaluate the robustness of BooST when pretrained with dynamic visual distractors. To generate a dataset with realistic visual noise, we augment LIBERO-90 by injecting an animatable human into the scene, as shown in Fig.~\ref{fig:distractor_libero_figure}. The human model is implemented in MuJoCo via MJCF schema~\cite{todorov2012mujoco} and parameterized SMPL~\cite{SMPL:2015}, which provides a kinematic joint hierarchy and a deformable surface mesh. For each episode, a single human distractor is instantiated by sampling one of 23 human texture maps~\cite{casas2023smplitex} and one of 23 motion sequences from AMASS~\cite{AMASS:ICCV:2019}, followed by random placement and scaling within the workspace. These distractors introduce task-irrelevant visual motion, making the skill learning process more challenging. We pretrain BooST and all baselines on this \emph{distractor-augmented} LIBERO-90 dataset and then evaluate downstream adaptation on the \emph{standard} LIBERO-90, Goal, Object, and Spatial benchmarks, each containing 50 demonstrations per task. This setup assesses whether the pretrained skill representations remain robust and transferable when exposed to dynamic and semantically irrelevant visual variations.

\noindent\textbf{Baselines.} We compare BooST with latent action pretraining frameworks that learn a discrete latent space from large-scale vision-language data, similar to our approach. Specifically, we include LAPA~\cite{ye2024latent} and UniVLA~\cite{bu2025univla}. These methods learn latent actions from unlabeled videos by modeling visual transitions via inverse or forward dynamics, thereby embedding motion information into discrete latent representations. Given limited computational resources, we adopt their latent action pretraining strategies while keeping the skill prior and low-level policy architectures identical to those of BooST. This setup allows us to isolate the effect of latent pretraining objectives on robustness under dynamic visual distractors.

\begin{table}[t]
  \centering
  \caption{\textbf{Performance on LIBERO benchmarks with dynamic visual distractors.} We report average success rates over three random seeds, each evaluated across five rollouts per seed. \textbf{Bold} denotes the best result per column.}
  \label{tab:distractor_libero_performance}
  \small
  
  \renewcommand{\arraystretch}{1.2}
  \begin{tabularx}{\columnwidth}{ l *{4}{>{\centering\arraybackslash}X} >{\centering\arraybackslash}X }
    \toprule
    {\textbf{Method}} & \textbf{90} & \textbf{Goal} & \textbf{Object} & \textbf{Spatial} & \textbf{Avg.} \\
    \midrule
    LAPA~\cite{ye2024latent} 
      & \meanstd{0.68}{0.04} 
      & \meanstd{0.69}{0.08} 
      & \meanstd{0.91}{0.08} 
      & \meanstd{0.87}{0.06} 
      & \meanstd{0.79}{0.03} \\
      
    UniVLA~\cite{bu2025univla} 
      & \meanstd{0.62}{0.03} 
      & \meanstd{0.49}{0.10} 
      & \meanstd{0.90}{0.05} 
      & \meanstd{0.80}{0.08} 
      & \meanstd{0.70}{0.01} \\
      
    \midrule
    \rowcolor{ourscellbg}
    \textbf{BooST (Ours)} 
      & \bmeanstd{0.89}{0.01} 
      & \bmeanstd{0.88}{0.05} 
      & \bmeanstd{0.97}{0.03} 
      & \bmeanstd{0.88}{0.05} 
      & \bmeanstd{0.90}{0.01} \\
    \bottomrule
  \end{tabularx}
  \vspace{-5mm}
\end{table}

\begin{figure}[!t]
  \centering 
  \includegraphics[width=1.0\columnwidth]{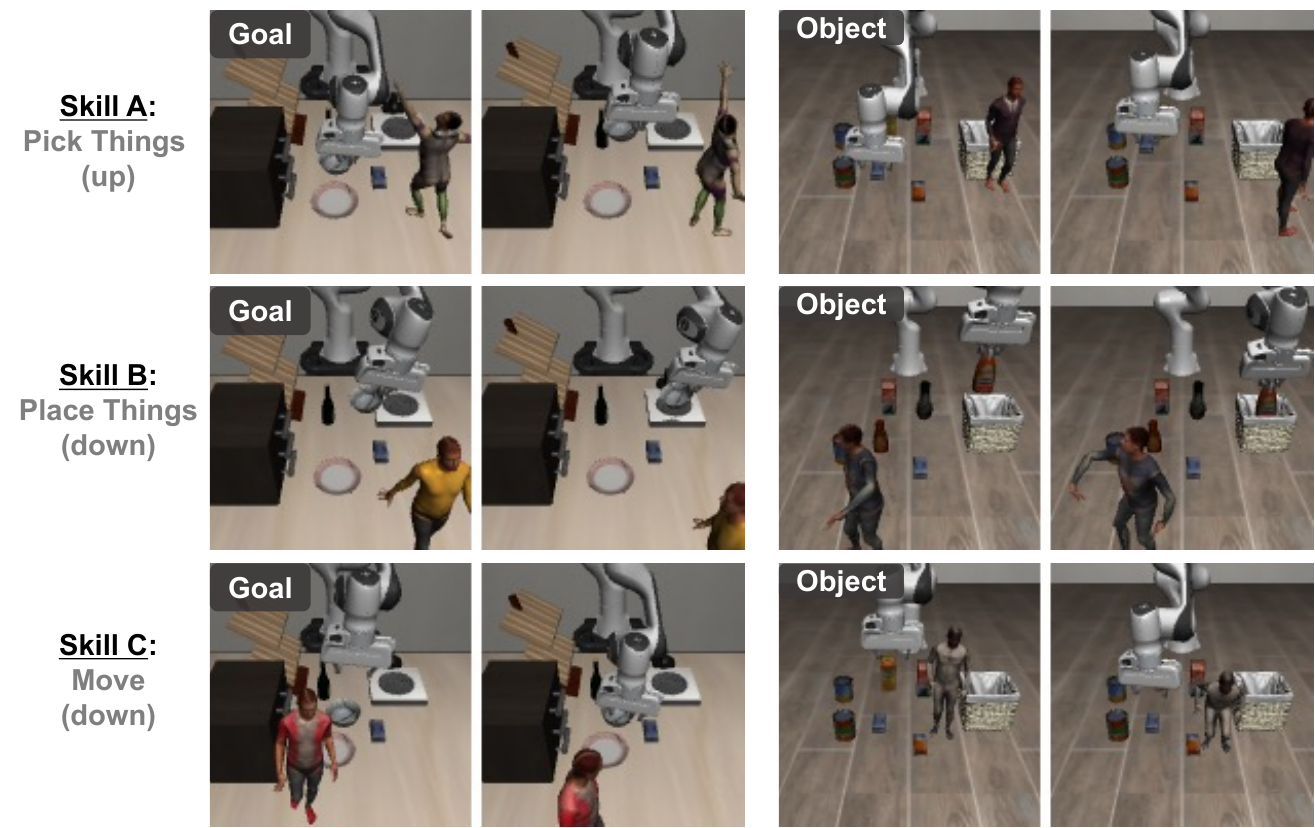}
  \caption{\textbf{Visualization of skills under test-time dynamic distractors.} The model is pretrained on distractor-augmented LIBERO-90 and evaluated on LIBERO-Goal and Object with dynamic human distractors present at test time. Despite the presence of dynamic background motion, each skill preserves consistent semantic intent and motion dynamics.}
  \label{fig:distractor_libero_figure}
  \vspace{-5mm}
\end{figure}

\noindent\textbf{Results.} Table~\ref{tab:distractor_libero_performance} shows that BooST maintains high adaptation performance even when pretrained on distractor-augmented data, outperforming latent action baselines.
Both LAPA and UniVLA exhibit poor performance under distractor conditions. 
Since their latent action learning reconstructs images, the representations inadvertently encode background motion and task-irrelevant visual variations, rather than the agent's own behavior. 
As a result, their learned representations are sensitive to moving distractors and fail to maintain consistent motion dynamics. In contrast, BooST explicitly grounds skill learning in action reconstruction and cross-modal alignment, and further extracts task-aware visual features, collectively forcing the model to focus on task-relevant behavior instead of external noise. Consequently, BooST's unified skill captures what the agent \emph{did}, rather than what merely \emph{moved} in the scene. 
Fig.~\ref{fig:distractor_libero_figure} illustrates the robustness at both training and test time. The skills are learned from distractor-augmented LIBERO-90 and visualized on \emph{unseen} benchmarks (Goal, Object) that also contain unseen dynamic distractors. The skill encoder consistently selects the same skill for a given sub-behavior across scenes, despite distractors during both pretraining and evaluation.

\subsection{Ablation Studies}
We conduct two ablation studies on the contribution of each component and the efficiency of our design.
First, we remove the \emph{action pathway} (\(E_{\text{act}}\)) and the \emph{task-aware encoder} (\(E_{\text{task}}\)) during skill pretraining.
In the latter case, the pretrained CLIP encoder and its language-conditioned task-aware visual feature extraction are replaced with a ResNet-34 trained from scratch. All variants are pretrained same as Sec.~\ref{sec:experiment_simulation}, except the number of demonstrations is fixed at 10 per task. As shown in Table~\ref{tab:ablation_studies}, both modules are crucial for achieving strong adaptation performance. The absence of \(E_{\text{act}}\) impairs motion dynamics. Although the visuo-linguistic pathway still reconstructs actions---thereby retaining a coarse motion signal---this lack of explicit action encoding compromises motion fidelity. Separately, omitting \(E_{\text{task}}\) weakens semantic grounding and generalization, leading to poorer adaptation.
Second, we vary the number of parameters in the downstream model---comprising the skill prior (\(p_{\psi}\)) and low-level policy (\(\pi_{\theta}\))---and measure downstream success rate. As shown in Table~\ref{tab:model_size_comparison}, smaller configurations retain competitive performance, suggesting that BooST's unified skill representation supplies the expressiveness needed for control even with lightweight models.
Across all configurations, the downstream models run at approximately 60 Hz, indicating inference efficiency suitable for real-robot deployment. For reference, our implementations of Diffusion Policy, VQ-BeT, and QueST run at 12 Hz, 95 Hz, and 30 Hz, respectively.

\begin{table}[t]
  \centering
  \caption{\textbf{Ablation studies}. Success rates are averaged over three seeds (five rollouts per seed), using 10 demonstrations for each task. \textbf{Bold} indicates the best performance in each column.}
  \label{tab:ablation_studies}

  \footnotesize
  
  \setlength{\tabcolsep}{3pt} 
    
  \begin{tabularx}{\columnwidth}{ l *{4}{>{\centering\arraybackslash}X} }
    \toprule
    \textbf{Ablated Components} & \textbf{90} & \textbf{Goal} & \textbf{Object} & \textbf{Spatial} \\
    \midrule
    w/o. Action Pathway (\(E_{\text{act}}\)) 
      & \meanstd{0.57}{0.07} 
      & \meanstd{0.57}{0.11} 
      & \meanstd{0.67}{0.04} 
      & \meanstd{0.45}{0.05} \\
      
    w/o. Task-Aware Encoder (\(E_{\text{task}}\)) 
      & \meanstd{0.25}{0.02} 
      & \meanstd{0.14}{0.02} 
      & \meanstd{0.55}{0.04} 
      & \meanstd{0.30}{0.03} \\
      
    \midrule
    \rowcolor{ourscellbg}
    \textbf{BooST (Ours)} 
      & \bmeanstd{0.70}{0.02} 
      & \bmeanstd{0.67}{0.05} 
      & \bmeanstd{0.78}{0.07} 
      & \bmeanstd{0.59}{0.08} \\
    \bottomrule
  \end{tabularx}
  \vspace{-1mm}
\end{table}

\begin{table}[t]
  \centering
  \caption{\textbf{Comparison of model size and performance.} Success rates are averaged over three seeds (five rollouts per seed). \textbf{Bold} indicates the best performance in each column.}
  \label{tab:model_size_comparison}
  
  \footnotesize 
  \renewcommand{\arraystretch}{1.0} 
  
  \begin{tabularx}{\columnwidth}{ c | *{4}{>{\centering\arraybackslash}X} } 
    \toprule
    
    \textbf{Model Size} & 
    \textbf{Goal} & 
    \textbf{Object} & 
    \textbf{Spatial} &
    \textbf{Avg.} \\
    
    \midrule
    29.7M 
      & \meanstd{0.93}{0.04} 
      & \meanstd{0.91}{0.06} 
      & \meanstd{0.83}{0.01} 
      & \meanstd{0.89}{0.02} \\
      
    69.5M 
      & \meanstd{0.93}{0.01} 
      & \bmeanstd{0.97}{0.01} 
      & \bmeanstd{0.91}{0.05} 
      & \bmeanstd{0.94}{0.02} \\
      
    144.5M 
      & \bmeanstd{0.96}{0.02} 
      & \meanstd{0.93}{0.05} 
      & \bmeanstd{0.91}{0.04} 
      & \meanstd{0.93}{0.02} \\
      
    \bottomrule
  \end{tabularx}
  \vspace{-4mm}
\end{table}

\subsection{Limitations}
Despite the strong performance, there remain opportunities for further improvement. First, because skills are extracted from 2D images, performance can degrade on tasks that demand especially precise motion along the z-axis in camera coordinates. Second, to enable lightweight deployment, the distilled skill prior is implemented with a non-CLIP-based network, which can face challenges under large viewpoint changes that require stronger 3D understanding.

\section{Conclusions}
Our work introduces a two-stage framework that learns a unified skill representation by jointly capturing semantic intent and motion dynamics, and then distills this representation into a lightweight policy and skill prior for efficient adaptation. Extensive experiments in simulation and real-world settings demonstrate that BooST enables strong few-shot adaptation, cross-embodiment skill transfer, and robustness to dynamic visual distractors and background variations. These results indicate that BooST attains the three desiderata for efficient skill transfer—generalization, robustness, and efficiency. 
A promising future direction is to extend BooST with 3D-aware representations, for instance by incorporating depth estimation models~\cite{yang2024depth} (e.g., DepthAnything~\cite{yang2024depth}), to enable more fine-grained 3D skill extraction.

\bibliographystyle{IEEEtran}
\bibliography{ref}

\end{document}